\documentclass[nohyperref]{article}

\usepackage{microtype}
\usepackage{graphicx}
\usepackage{subfigure}
\usepackage{booktabs} 

\usepackage{hyperref}

\usepackage[accepted]{icml2022}

\usepackage{amsmath}
\usepackage{amssymb}
\usepackage{mathtools}
\usepackage{amsthm}
\usepackage{bm}
\usepackage[normalem]{ulem}
\usepackage{comment}
\usepackage{url}

\usepackage[capitalize,noabbrev]{cleveref}
\newcommand{\ph}[1]{\vspace{1mm} \noindent \textbf{#1.}\quad}

\theoremstyle{plain}

\theoremstyle{definition}

\theoremstyle{remark}

\usepackage[textsize=tiny]{todonotes}

\newcommand{\method}{{Gate-Drop}}
\newcommand{\extension}{{Gate-Expert-Drop}}
\newcommand{\extensiontwo}{{Gate-Expert-Drop}}
\icmltitlerunning{Gating Dropout: Communication-efficient Regularization for Sparsely Activated Transformers}

\begin{document}

\twocolumn[
\icmltitle{Gating Dropout: Communication-efficient Regularization for\\ Sparsely Activated Transformers}



\icmlsetsymbol{intern}{*}




\begin{icmlauthorlist}
\icmlauthor{Rui Liu}{intern,umich}
\icmlauthor{Young Jin Kim}{ms}
\icmlauthor{Alexandre Muzio}{ms}
\icmlauthor{Hany Hassan Awadalla}{ms}
\end{icmlauthorlist}

\icmlaffiliation{umich}{University of Michigan, Ann Arbor}
\icmlaffiliation{ms}{Microsoft}

\icmlcorrespondingauthor{Rui Liu}{ruixliu@umich.edu}
\icmlcorrespondingauthor{Young Jin Kim}{youki@microsoft.com}

\icmlkeywords{Machine Learning, ICML}

\vskip 0.3in
]



\printAffiliationsAndNotice{\textsuperscript{*}Work done while a research intern at Microsoft}  

\begin{abstract}
Sparsely activated transformers, such as Mixture of Experts (MoE), have received great interest due to their outrageous scaling capability which enables dramatical increases in model size without significant increases in computational cost. To achieve this, MoE models replace the feedforward sub-layer with Mixture-of-Experts sub-layer in transformers and use a gating network to route each token to its assigned experts. Since the common practice for efficient training of such models requires distributing experts and tokens across different machines, this routing strategy often incurs huge cross-machine communication cost because tokens and their assigned experts likely reside in different machines. In this paper, we propose \emph{Gating Dropout}, which allows tokens to ignore the gating network and stay at their local machines, thus reducing the cross-machine communication. 
Similar to traditional dropout, we also show that Gating Dropout has a regularization effect during training, resulting in improved generalization performance. We validate the effectiveness of Gating Dropout on multilingual machine translation tasks. Our results demonstrate that Gating Dropout improves a state-of-the-art MoE model~\cite{kim2021scalable} with faster wall-clock time convergence rates and better BLEU scores for a variety of model sizes and datasets.
\end{abstract}

\section{Introduction}
Large transformer models have been shown to be effective and powerful in many natural language processing tasks such as machine translation~\cite{lewis2019bart,conneau2019cross} and also language understanding~\cite{devlin2018bert,liu2019roberta,brown2020language,radford2019language}.
Sustaining the continued growth in size for these dense models requires tremendous compute resources, since all the parameters are used for all input examples. 
Seeking better computational efficiency, sparsely activated models have gained  great interest; since they utilize  a subset of the model weights for each input example~\cite{shazeer2017outrageously,lepikhin2020gshard}.
This leads to a desirable characteristic: sparsely activated models can scale up  to an order of magnitude larger than dense models  without significant increase in computational cost.
However, libraries and hardware accelerators widely used by machine learning practitioners are much efficient in supporting dense matrix operations. 
Mixture of Experts (MoE) models are a specific type of sparsely activated models that have had notable success in machine translation~\cite{shazeer2018mesh,lepikhin2020gshard} and recently been explored in other tasks~\cite{riquelme2021scaling} while running efficiently on top of existing libraries and hardware accelerators.
In this paper, we focus on 
multilingual machine translation tasks. 

A dense transformer model consists of a stack of layers composed of two sub-layers: a multi-head self-attention sub-layer followed by a feedforward network (FFN) sub-layer.
An MoE model replaces the FFN sub-layer with a Mixture-of-Experts (MoE) sub-layer which takes as input a token representation and routes this token to the best top-$k$ experts. 
The MoE sub-layer is essentially a group of experts with each expert still a FFN sub-layer. 
Typically, $k$ is a  very small value.
As with~\cite{fedus2021switch,kim2021scalable}, we assume $k=1$ by default.
Note that our method can also be extended to the case when $k>1$.
During training and inference, an input token is routed to only one expert ($k=1$), resulting in constant computational cost for one forward pass regardless of the total number of experts.
The target expert for each token is determined by a gating network which is typically a small one-layer FFN.

To achieve highly efficient training, a combination of data, expert, and model parallelism is adopted: (1) the part of the model excluding MoE sub-layers is replicated across multiple machines in a data-parallel fashion, (2) experts from each MoE sub-layer are split across machines (i.e., different machines hold the parameters of different experts), and (3) techniques such as tensor slicing\footnote{To understand the key idea of Gating Dropout, readers can focus on data and expert parallelism (i.e., ignore model parallelism such as tensor slicing) to simplify the complexity of parallelism.} are used to partition tensors among multiple machines if they are too large for a single machine (e.g., partition along the $d_{ff}$ dimension inside each expert when $d_{ff}$ is too large)~\cite{fedus2021switch}. 
Since tokens and their assigned experts are likely to reside at different machines, we need to shuffle the tokens around via the communication link among machines. 
This is typically achieved by using the all-to-all collective operation, which is very costly even if the data is cast to bfloat16 precision~\cite{fedus2021switch}.
If we assume that token representation is of $d$ dimension, the sequence length is $L$, total batch size is $B$, then the total amount of data that needs to be handled by the all-to-all operation is $2BLd$ bytes\footnote{The factor $2$ comes from the bfloat16 data representation, which uses 2 bytes per number.}.
As a commonly used setting in practice, we assume $d=4096=2^{12}, L=1024=2^{10}, B=128=2^7$, then $2BLd= 2^{1+12+10+7} = 2^{30} = 1G$.
In other words, the all-to-all operation  needs to shuffle  roughly $1GB$ of data at each MoE sub-layer for each forward pass. Therefore, the communication cost from all-to-all operations is greatly affecting the training speed for MoE models.

In this paper, we propose \emph{Gating Dropout} as a communication-efficient regularization technique for training sparsely activated transformers, specifically for MoE models.
Gating Dropout allows tokens to ignore the assignments from the gating network and stay at their local machines with certain probability, thus reducing the need for routing the tokens to experts at different machines.
This significantly improves the throughput and convergence speed for MoE training. 
In addition, by forcing tokens to be routed to experts on the same machines (which might not be the optimal experts), we encourage the experts to learn a general ability.
We think this leads to a regularization effect during training, resulting in better generalization performance when training is converged.

To validate the effectiveness of Gating Dropout, we have conducted extensive experiments on multilingual machine translation tasks using different model settings. With Gating Dropout, our results show that we improve a state-of-the-art MoE model~\cite{kim2021scalable} in terms of convergence speed (up to $58\%$ less training time to target BLEU score), generalization performance (up to $0.6$ BLEU score increase) and throughput (up to $16\%$ more tokens per second). Our code will be available at \url{https://aka.ms/gating_dropout}.


\begin{figure*}
\hspace{0.3cm}
\begin{minipage}{0.3\textwidth}
\vspace{1.3cm}
    \centering
    \includegraphics[width=4 cm]{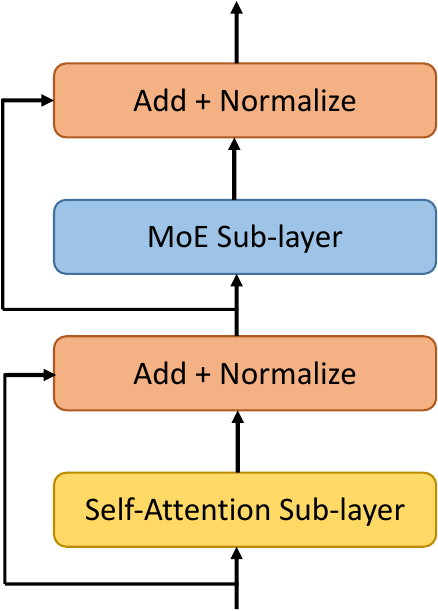}
    \vspace{0.8cm}
    \caption{The MoE model architecture. The FFN sub-layer in the original transformer architecture is replaced by the MoE sub-layer, which consists of experts.}
    \label{fig:moe_arch}
\end{minipage}
\hspace{0.15cm}
\begin{minipage}{0.65\textwidth}
    \centering
    \includegraphics[width=10 cm]{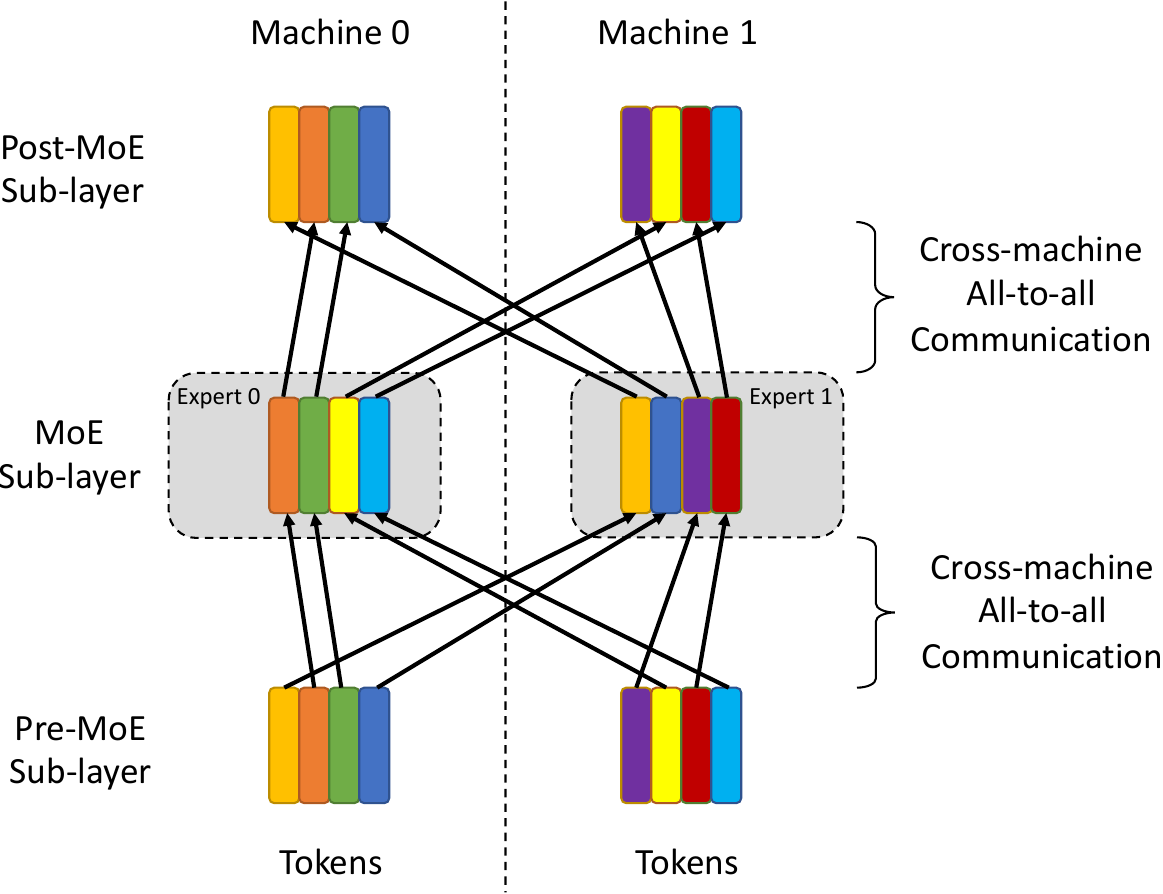}
    \caption{The all-to-all communication illustration of an MoE model with two experts in a two-machine setting. Expert 0 and 1 reside at machine 0 and 1, respectively. Each machine has a local batch of $4$ tokens to process. All-to-all communication is needed right before (and right after) MoE sub-layer.}
    \label{fig:moe_alltoall_diagram}
\end{minipage}
\end{figure*}

\section{Background and Motivation}
\subsection{From Densely Activated Transformers to Sparsely Activated Transformers}

In~\citet{vaswani2017attention}, the transformer architecture was introduced to the NLP community due to their superior performance in sequence-to-sequence tasks, such as neural machine translation. 
A sequence-to-sequence transformer contains an encoder module and a decoder module. 
The encoder module is a stack of encoder layers, each employing a self-attention sub-layer followed by an FFN sub-layer. 
The decoder module is a stack of decoder layers, each employing a self-attention sub-layer followed by a cross-attention sub-layer with an FFN sub-layer at the end.
Ever since, transformer-based models have been the top performers in various NLP tasks, such as BERT~\cite{devlin2018bert}, RoBERTa~\cite{liu2019roberta}, GPT-3~\cite{brown2020language}, XLNet~\cite{yang2019xlnet}, ELECTRA~\cite{clark2020electra} and TMR~\cite{liu2022transformer}.
However, these transformer models are densely activated, meaning that all model parameters are used to process all input examples. 
Scaling up the model size for them results in significant increase in computational cost during both training and inference. 
Therefore, the level of scaling depends on the availability of compute resources (e.g., GPU or TPU), which can be costly to obtain. 
For example, it is estimated that it takes $168$ days to train a GPT-3 model with $178$ billion parameters using $256$ NVIDIA A100 GPUs~\cite{narayanan2021efficient}. 
Megatron-Turing NLG $530$B model with $530$ billion parameters
had been trained  on NVIDIA’s Selene  supercomputer
with $560$ DGX A100 nodes, each node having $8$ NVIDIA $80$GB A100 GPUs connected to each other with high bandwidth Infiniband configuration~\cite{megatron-turring}.

On the other hand, sparsely activated transformers use a subset of model parameters for each different input example, leading to outrageous scaling capability.
An Mixture-of-Experts (MoE)~\cite{fedus2021switch} model has been scaled to $1.5$ trillion parameters while GPT-3, one of the largest dense models, contains only $175$ billion parameters. 
The core component of MoE models is the Mixture-of-Experts (MoE) sub-layer, which replaces the FFN sub-layer in the dense transformers, as illustrated in Figure~\ref{fig:moe_arch}. 
Let $\{E_i\}_{i=1}^N$ denote the set of experts at an MoE sub-layer where $N$ is the total number of experts. 
Each expert $E_i$ is still a feedforward network (FFN).
The parameters of one expert is independent from those of another expert, meaning that they can be trained to converge to different values.
Each token is routed to the top-$k$ experts, which are determined by a gating network. 
Given a token $x\in \mathbb{R}^d$, where $d$ is the hidden dimension, the output of a gating network is logits $h(x)=W_r\cdot x$, followed by a normalization via a softmax, i.e., 
\begin{equation}
    p_i(x) = \frac{e^{h(x)_i}}{\sum_{j=1}^N e^{h(x)_j}},
\end{equation}
where $W_r\in\mathbb{R}^{N\times d}$ is the trainable weight matrix of the gating network.
Given the gating network output $\{p_i(x)\}_{i=1}^N$, we select the top-$k$ experts to form the set of activated experts $\mathcal{T}\subset \{1, \cdots, N\}$, where $|\mathcal{T}|=k$. 
In other words, the token $x$ is assigned to experts in $\mathcal{T}$.
Typically $k$ is much smaller than $N$. 
In~\citet{fedus2021switch}, it is shown that using $k=1$ is sufficient with better computational efficiency. 
In this paper, we assume $k=1$ by default.
Then the output of an MoE sub-layer for the input token $x$ is 
\begin{equation}
    y=\sum_{i\in \mathcal{T}} p_i(x)E_i(x).
\end{equation}

\begin{figure}
    \centering
    \includegraphics[width=6 cm]{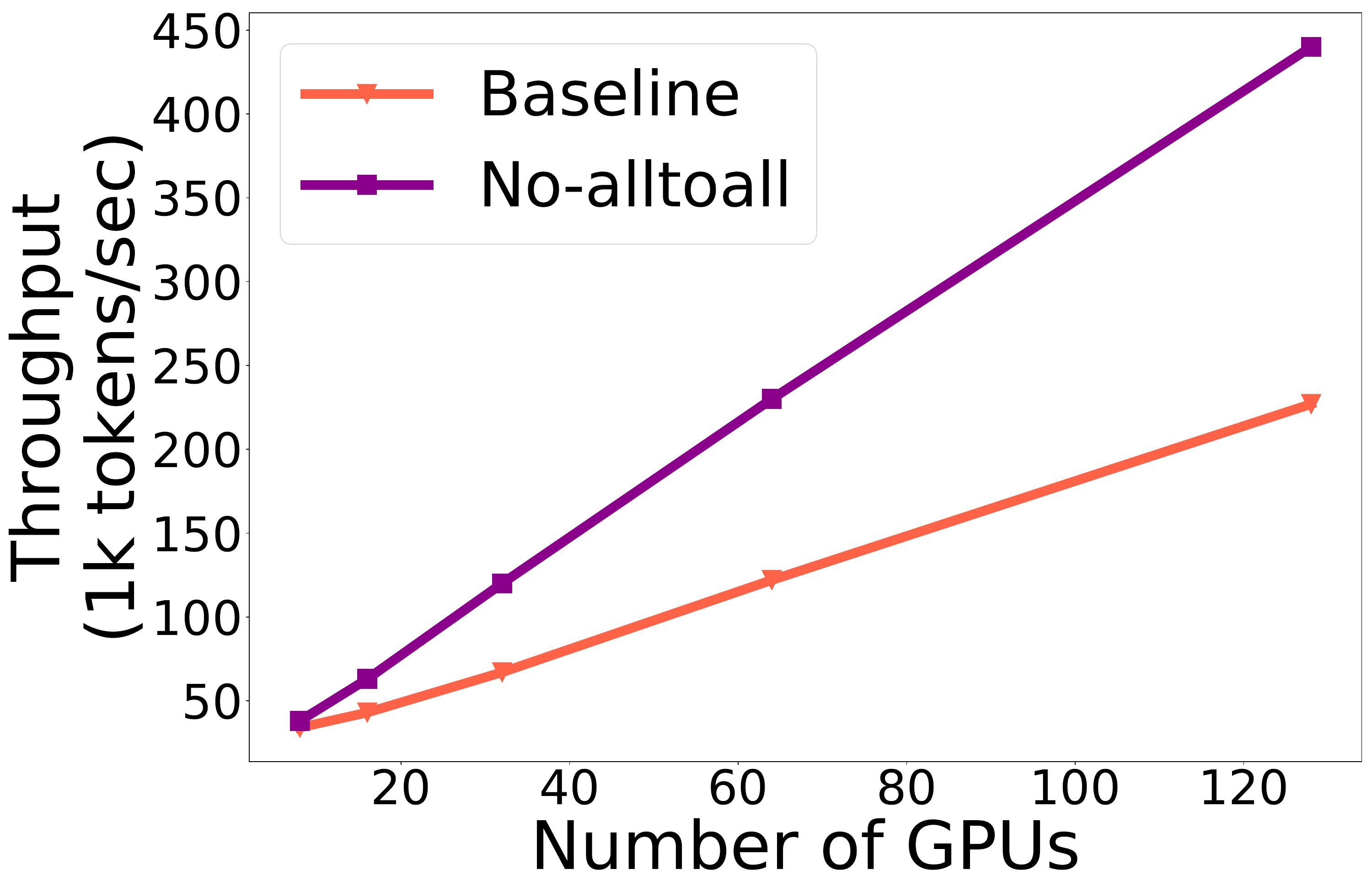}
    \caption{The all-to-all communication cost.}
    \label{fig:moe_alltoall_cost}
\end{figure}

\begin{table}[t]
\caption{Relative throughput improvement of no-alltoall compared to baseline.}
\label{table:moe_alltoall_increase}
\begin{center}
\begin{tabular}{cc}
\hline
\multicolumn{1}{c}{\bf Number of GPUs} & \multicolumn{1}{c}{\bf Throughput Impr.}\\
\hline
 $8$ &$11.8\%$ \\
 $16$ & $46.5\%$ \\
 $32$ & $79.1\%$ \\
 $64$  & $88.5\%$ \\
 $128$ & $93.8\%$ \\
\hline
\end{tabular}
\end{center}
\vspace{-0.5cm}
\end{table}

\subsection{All-to-all Communication Cost of MoE Models}
Distributed training involving multiple machines is typically used to support highly scaled MoE models. 
Each machine contains a replication of the model parameters excluding experts (i.e., data parallelism), and a subset of experts (i.e., expert parallelism). 
Since experts are split across multiple machines, each token needs to be sent to the machine where its assigned expert resides.
This is typically achieved by using the all-to-all collective communication~\cite{fedus2021switch}, which is illustrated in Figure~\ref{fig:moe_alltoall_diagram} using a two-machine setting.
We conducted some experiments to demonstrate the all-to-all communication cost (see the settings on WMT-10 dataset in Section~\ref{sec:exp_setup} for details).
Specifically, we use a cluster of NVIDIA V100 GPUs connected via a $100$Gb/s InfiniBand fabric.
We use a state-of-the-art MoE model~\cite{kim2021scalable} as the baseline, and a no-alltoall variant which is the same as baseline but with  the all-to-all operations entirely skipped. 
We compare the throughput in terms of the number of tokens processed per second between the baseline and the no-alltoall variant. 
To see how the throughput changes as we scale up the model, we increase the number of GPUs from 8 all the way to 128, and set the number of experts the same as the number of GPUs due to expert parallelism~\cite{fedus2021switch}.
From Figure~\ref{fig:moe_alltoall_cost}, we can see that as we increase the number of GPUs, the throughput for both the baseline model and no-alltoall variant increases. 
This is expected because using more GPUs provides greater computational power which increase the processing speed.
We also observe that the no-alltoall variant has significantly higher throughput than the baseline when using the same number of GPUs. 
In Table~\ref{table:moe_alltoall_increase}, we list the relative throughput improvement of the no-alltoall compared to the baseline.
As we increase the number of GPUs, the relative throughput improvement also increases.
Especially, when the number of GPUs is $128$, the relative improvement is more than $90\%$.
Note that the difference between the baseline and no-alltoall comes from whether the all-to-all communication in the MoE sub-layers is invoked or not.
From these results, we can see that all-to-all communication greatly affects the training speed.
As the number of GPUs increases, the all-to-all communication becomes more costly because data exchange involves more machines 
and communication cost is proportional to the number of involved machines. 
In summary, to better support the scaling capability of MoE models, it will pay off to investigate methods that can reduce the all-to-all communication cost without sacrificing the performance.

\begin{figure*}
    \centering
    \includegraphics[width=13.5 cm]{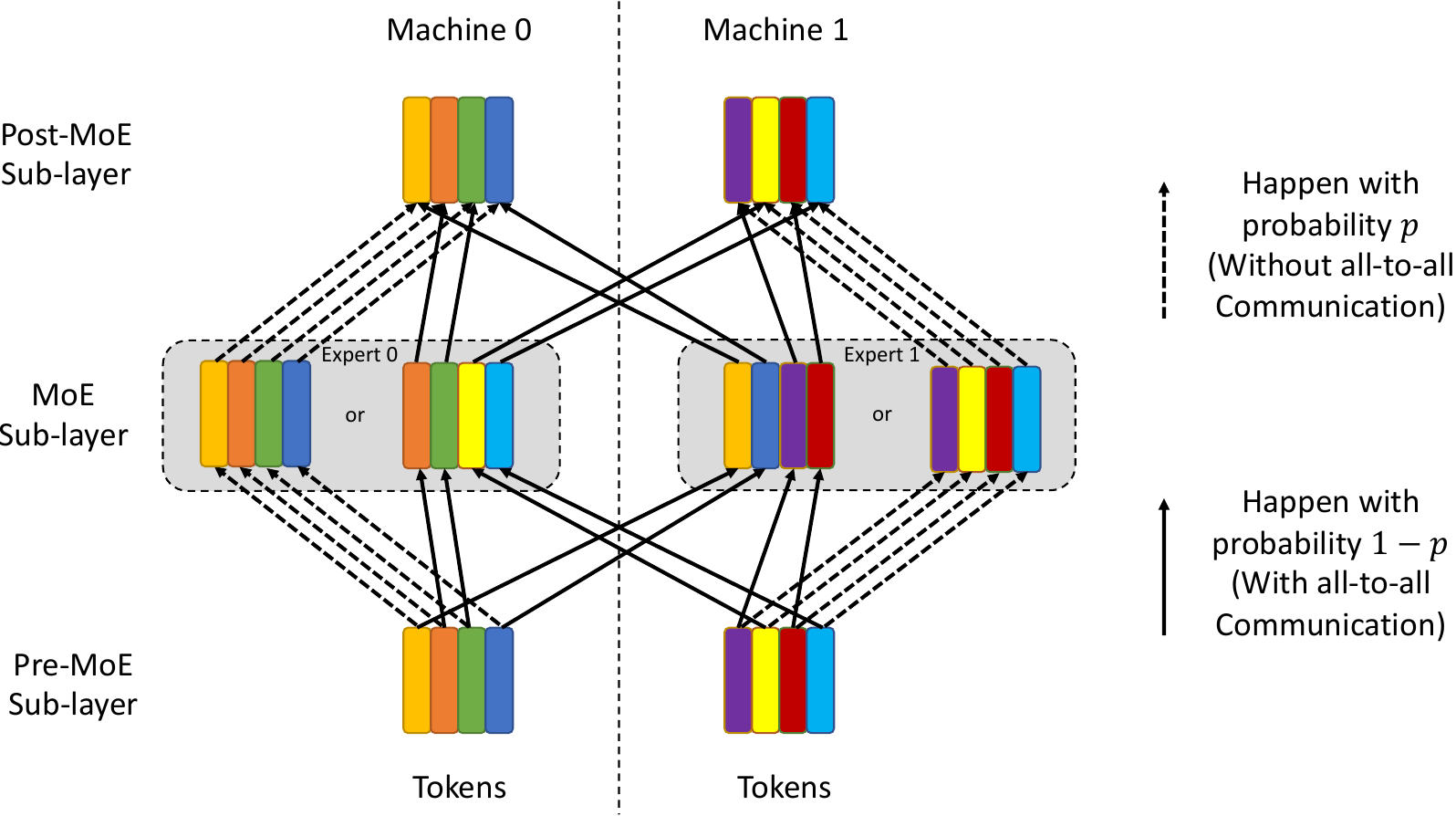}
    \caption{The Gating Dropout illustration of an MoE model with two experts in a two-machine setting. Expert 0 and 1 reside at machine 0 and 1, respectively. Each machine has a local batch of $4$ tokens to process. Gating Dropout allows us to send tokens to the local experts at the same machines with probability $p$ (denoted by the dashed arrows).}
    \label{fig:moe_gating_dropout_diagram}
\end{figure*}

\section{Gating Dropout as Communication-efficient Regularization}
In this section, we describe \emph{Gating Dropout} as an effective way to reduce the all-to-all communication cost. 
We also show that Gating Dropout has a regularization effect during training, resulting in improved generalization performance. 
Gating Dropout  essentially combines communication efficiency with regularization, which has been challenging to achieve yet desirable  in distributed training, where communication cost is often the bottleneck.
Gating Dropout works in the following way, which is also illustrated in Figure~\ref{fig:moe_gating_dropout_diagram}, at each training iteration:
\begin{itemize}
    \item with probability $p$, all the tokens are sent to the local experts at the same machine without all-to-all communication involved;
    \item with probability $1-p$, all the tokens are sent to their assigned experts with all-to-all communication involved, as is done in normal MoE models. 
\end{itemize}
The above probability $p$ is the dropout rate. Note that, since all-to-all is a collective operation requiring all the machines to simultaneously participate, the decision of turning on/off Gating Dropout should be consensual among all the machines at each iteration.
In other words, at each iteration, it is either (1) that all the machines keep their tokens locally (i.e., Gating Dropout is turned on), or (2) that  all the machines use the all-to-all operations to send tokens to their assigned experts (i.e., Gating Dropout is turned off). 
To make sure all machines achieve consensual decisions, we appoint one machine as the coordinator, responsible for making the randomized decision, and broadcasting the decision to all the machines at each iteration.
The overhead of broadcasting the decision is negligible, because the decision can be represented by a binary value.

During inference, we simply turn off the Gating Dropout by setting $p=0$. We do not have the additional weight scaling step as in the traditional dropout (i.e., multiply the weight by $p$). In the traditional dropout, the weight scaling step is used to ensure the expected neuron output is the same between inference time and training time, because traditional dropout works by setting the neuron output as $0$. However, Gating Dropout will not affect expected neuron output because it works by modifying the token routing strategy.

This technique can be easily implemented with several lines of code, by using the broadcast collective operation for broadcasting the decision and the conditional branch for skipping the all-to-all communication. 
We refer to this basic version of Gating Dropout as \method{} in order to differentiate it from the extended version that will be introduced in Section~\ref{sec:expert_drop}.

The benefits of Gating Dropout are twofold:
\begin{itemize}
    \item higher throughput due to reduced communication cost;
    \item better generalization performance due to its regularization effect.
\end{itemize}

\ph{Communication Efficiency}
Gating Dropout reduces the all-to-all communication cost, thus improving the throughput during training. 
When the Gating Dropout is turned on at a certain iteration, all the tokens go to the local experts at the same machines. 
In this case, all-to-all communication is no longer needed at this iteration. 
However, when Gating Dropout is turned off, the all-to-all communication is still used.
The dropout rate $p$ essentially controls how often the gating dropout is turned-on on average, thus affecting the throughput during the training process.
The no-alltoall variant in Figure~\ref{fig:moe_alltoall_cost} corresponds to the case when $p=1$, which provides the upper-bound of the  throughput improvement. It is worth noting that it is not possible to achieve this upper-bound since the model will not be able to learn any gating in such settings.
Therefore, the model performance will deteriorate when $p$ is too large. 
Next, we will show that Gating Dropout can lead to better performance when $p$ is properly chosen.

\ph{Regularization}
Gating Dropout can be viewed as a regularization technique to improve the model's generalization performance.
The gating network is used to determine which expert a token should be sent to. 
The gating network is supposed to select the best expert for each token, which is unknown in general.
Therefore, the common practice in MoE models is that the gating network is initialized randomly, and being trained together with all the experts.
Intuitively, Gating Dropout can break up the co-training between gating network and experts, just as traditional dropout is used to break up the co-adaptation among neurons~\cite{srivastava2014dropout}. 
This forces experts to work in a robust way  without relying too much on the guidance from the gating network (e.g., by obtaining a general ability).
Another perspective that affects the dynamics of training with Gating Dropout comes from exploration-exploitation trade-off~\cite{auer2002finite,sutton2018reinforcement}.
The gating network provides a discrete decision over all the experts.
This decision may not be optimal simply because other experts could be under-explored.
The epsilon-greedy method~\cite{sutton2018reinforcement} is a simple yet effective way to handle the exploration-exploitaion trade-off: with probability $\epsilon$, make a random decision over all  candidate options; with probability $1 - \epsilon$, make the  greedily best decision based on the available information so far.
Gating Dropout can be viewed as a simplified variant of the epsilon-greedy method that is tailored for  MoE model training while accounting for  the communication cost.
Specifically, instead of choosing a random expert, we choose the local expert at the same machine because the all-to-all communication is not needed to send tokens to their local experts.
In our experiments, it is observed that this simplified variant of the epsilon-greedy method actually yields good performance improvement.

\ph{Gating Dropout vs Expert Dropout}
It is worth noting the differences between Gating and Expert dropout. Expert Dropout is recommended in~\citet{fedus2021switch}
as a technique to regularize the MoE models. 
However, both techniques have quite distinctive characteristics. 
Expert Dropout is used during fine-tuning on downstream tasks that have very few examples to overcome possible overfitting due to lack of enough examples.
However, Gating Dropout is used in pre-training.
Furthermore, Expert Dropout is essentially traditional dropout~\cite{srivastava2014dropout} where  the dropout rate inside MoE sub-layers is explicitly increased.
They observed better performance on several smaller downstream tasks by setting a smaller dropout rate at non-MoE sub-layers and a higher rate at MoE sub-layers.
On the other hand, Gating Dropout acts as a regularization technique that is completely different from traditional dropout.

\ph{Gating Dropout vs Other Exploration-Exploitation Techniques}
The exploration-exploitation trade-off is also considered in~\citet{fedus2021switch}. They investigated several techniques such as sample softmax, input dropout and input jitter. First, 
none of these techniques is communication efficient, which is one of the main benefits of our Gating Dropout. Second, Gating Dropout is compatible with these techniques, so that they can be complimentary used together during  training. For example, input jitter, which has the best empirical performance in their experiments, injects noise to the incoming token representation. 
Gating Dropout can  still be  applied regardless of  how the token representation is modified. 
In our experiments, we have adopted input jitter by default
in the baseline.


\subsection{Extension with Expert LayerDrop}\label{sec:expert_drop}
It has been shown that skipping certain layers of neural networks during training can reduce the training time and also has a regularization effect~\cite{huang2016deep,fan2019reducing}. 
Inspired by LayerDrop proposed in~\citet{fan2019reducing} which randomly drops layers in (densely activated) transformers,  
we consider an extension of Gating Dropout by dropping experts.
Specifically, when Gating Dropout is turned on, all the tokens skip the local experts and directly go to the sub-layer following the MoE sub-layer.
When the Gating Dropout is off, we still send the tokens to their assigned experts with all-to-all communication.
We call this extended version \extension{}, while the basic version from the previous section is referred to as \method{}.
This extension improves the two strengths of Gating Dropout, since (1) skipping experts reduces the computational cost, thus improving the throughput, and (2)  skipping layers randomly has been shown with a regularization effect.

\section{Experiments}
We evaluate our proposed Gating Dropout (both \method{} and \extension{}) and compare it against a state-of-the-art MoE model~\cite{kim2021scalable} on multilingual translation tasks. Multilingual large-scale translation tasks have  been a good use case for large dense and sparse transformer models due to their high level of complexity that necessitates such powerful models.
We use the Z-code M3 model from~\citet{kim2021scalable} as our \textbf{baseline}, which has been shown better than prior MoE models.
All of our implementations are based on DeepSpeed library\footnote{\url{https://github.com/microsoft/DeepSpeed}}, due to its strong support for distributed training.
We show that Gating Dropout can improve the throughput, BLEU score and convergence speed.

\begin{table*}[t]
\caption{The results on WMT-10 dataset using models with Transformer-base architecture. The target BLEU score is the value achieved by the baseline method at convergence, i.e., $24.60$. 
The training time to the target BLEU is in seconds. Hash-Layer uses a random hash function to replace the gating network~\cite{roller2021hash}. We use - in the table to indicate that Hash-Layer cannot reach the target BLEU.}
\label{table:small_scale_results}
\begin{center}
\begin{tabular}{ccccc}
\hline
\multicolumn{1}{c}{\bf Method} & \multicolumn{1}{c}{\bf Throughput} & \multicolumn{1}{c}{\bf BLEU@convergence}&\multicolumn{1}{c}{\bf Time to target BLEU}   & \multicolumn{1}{c}{\bf{Steps to target BLEU}} \\
\hline
 Baseline & $129$k &$24.60$  & $250$k & $74.24$k \\
 Hash-Layer & $135$k & $23.06$ & - & -\\
 \method{}& $143$k & $25.06$  &$147$k & $48.64$k \\
\extension{} & \bm{ $150$}{\bf k} & \bm{$25.12$} & \bm{$104$}{\bf k} & \bm{$35.84$}{\bf k}\\
\hline
\end{tabular}
\end{center}
\end{table*}

\subsection{Setup}\label{sec:exp_setup}
\ph{Datasets} Similar to prior work~\cite{kim2021scalable}, we use the following datasets in our experiments.

\begin{itemize}
    \item \textbf{WMT-10:} We use WMT-10 benchmark dataset which includes bitext data between English and other \textbf{10} languages~\cite{wang2020multi}. 
    There are $32.5$ million parallel sentences in total. 
    \item \textbf{Web-50:} We use a dataset composed of an in-house \textbf{web}-crawled parallel dataset augmented with data from CCNet~\cite{wenzek2019ccnet}. It contains $700$ million parallel sentences in \bm{$50$} different languages.
    This dataset is much bigger than the WMT-10 dataset.
\end{itemize}

\ph{Model Settings}
We use the settings recommended in~\cite{fedus2021switch} to set up the MoE models. 
Specifically, MoE sub-layers are added to replace every other FFN sub-layers in both the encoder and decoder of the transformer. 
The capacity factor is set to $1.0$ during training and $2.0$ during testing~\cite{fedus2021switch}.
Jittering noise is applied to the token representation right before the gating network.
An additional balancing loss with multiplicative coefficient $0.01$ is used to better balance the utilization among different experts.
We use two different model sizes for experiments on the above two datasets.
\begin{itemize}
\item \textbf{WMT-10:} We adopt the architectural settings according to Transformer-base architecture with 12 encoder layers and 6 decoder layers~\cite{vaswani2017attention}. For each MoE sub-layer, we use $128$ experts by default. The model contains around $5.6$ billion parameters.
\item \textbf{Web-50:} We adopt the architectural settings according to Transformer-big architecture with 24 encoder layers and 12 decoder layers~\cite{vaswani2017attention}. For each MoE sub-layer, we use $64$ experts by default. The model contains around $10$ billion parameters.
\end{itemize} 

\ph{Hardware}
We use a cluster of NVIDIA V100 GPUs connected via a $100$Gb/s InifiBand fabric in most experiments. For some experiments on the Web-50 dataset, we run jobs on a cluster of NVIDIA A100 GPUs connected via a $1.6$Tb/s InifiBand fabric. We use $16$ GPUs for WMT-10 and $64$ GPUs for Web-50 if not stated otherwise. 

\ph{Training Details}
For training, we use Adam as the optimizer with $\beta_1=0.9$ and $\beta=0.99$. The learning rate is set $0.03$ with $5000$ warm-up steps and inverse square root scheduler as proposed in~\citet{raffel2019exploring}. 
We set the batch size to be equivalent to $435$k tokens.
For the gating dropout rate $p$, we set it to $0.3$ for \method{} and $0.2$ for \extension{} by default, which are chosen because of their good performance (see Section~\ref{sec:dropout_rate}). 
For training on the Web-50 dataset, we use Denoising
Auto-Encoding (DAE) loss~\cite{vincent2008extracting} together with Machine Translation (MT) loss because multi-tasking learning has been shown to further improve the cross-language knowledge transfer~\cite{wang2020multi,kim2021scalable}.

\ph{Metrics}
We compare between different methods using the following three metrics:
\begin{itemize}
    \item \textbf{Throughput:} the number of tokens processed per second by the model during training.
    \item \textbf{Generalization Performance:} BLEU score on a holdout validation set. Especially, we report the case-sensitive detokenized sacreBLEU~\cite{post2018call}.
    \item \textbf{Convergence Speed:} the amount of training time to reach a target BLEU score  on a holdout validation set.
\end{itemize}

\begin{figure}
    \centering
    \includegraphics[width=7 cm]{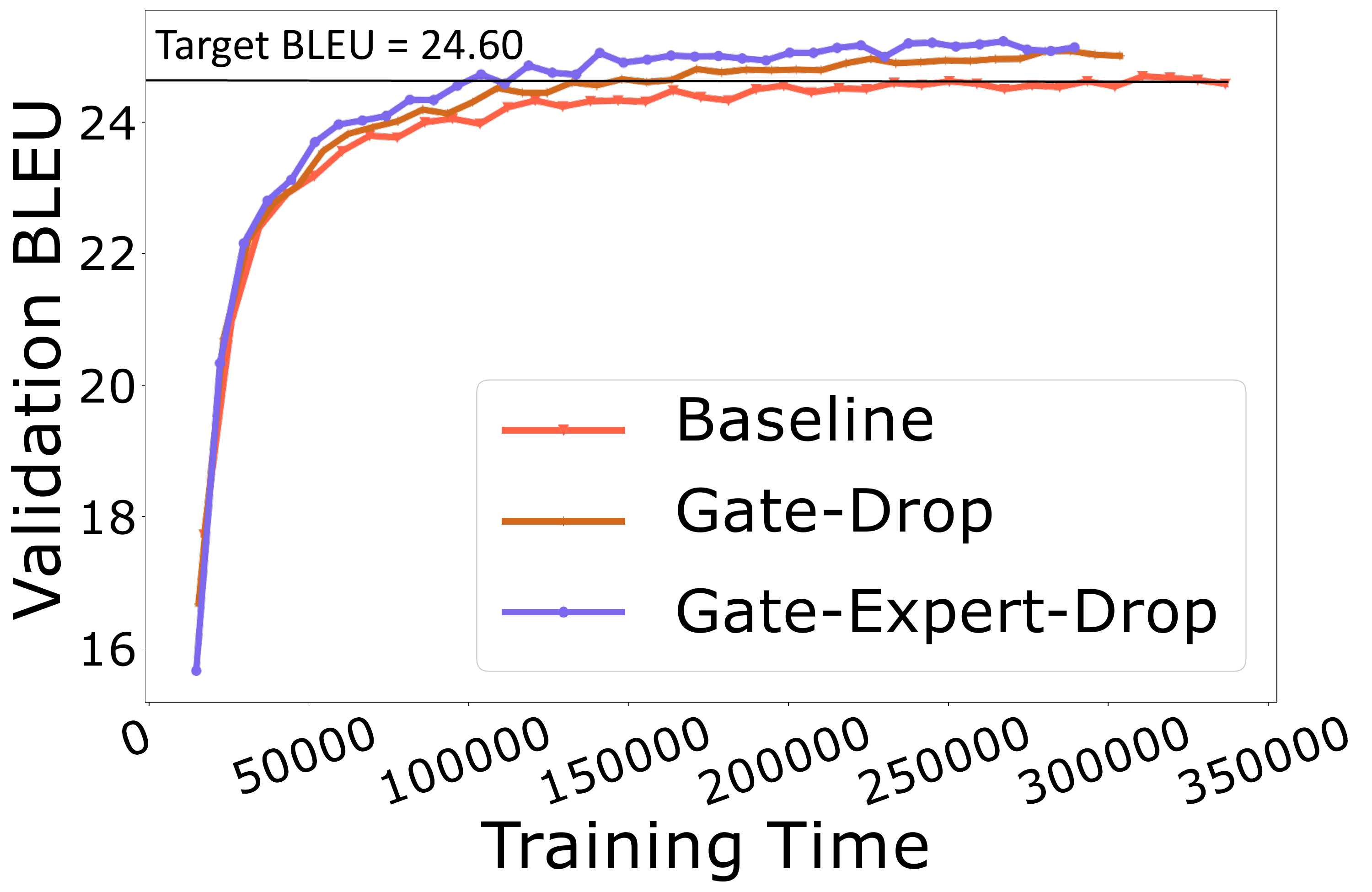}
    \caption{BLEU score vs. training time (in seconds) on the WMT-10 dataset. Both \method{} and \extension{} reach the target BLEU (denoted as the black line) much faster than the baseline.}
    \label{fig:small_bleu_curve}
\end{figure}

\begin{table}[t]
\caption{The throughput results on Web-50 dataset measured in two different clusters.}
\label{table:large_scale_throughput}
\begin{center}
\begin{tabular}{ccc}
\hline
\multicolumn{1}{c}{\bf Method} & \multicolumn{1}{c}{\bf V100 Cluster} & \multicolumn{1}{c}{\bf A100 Cluster} \\
\hline
 Baseline & $126$k & $362$k  \\
 \method{}&  $140$k &  $372$k \\
\begin{tabular}{@{}c@{}}\extensiontwo{}\end{tabular} &$146$k &  $384$k \\
\hline
\end{tabular}
\end{center}
\end{table}

\begin{table*}[t]
\caption{The results on Web-50 dataset using models with Transformer-big architecture. {\bf BLEU (avg)} is the average BLEU score for translating between English and other languages (both directions). {\bf E$\rightarrow$X} only considers translating from English to other languages. {\bf X$\rightarrow$E} only considers translating from other languages to English. And {\bf (low)} is used to indicate that only low-resource languages are considered.  }
\label{table:large_scale_bleu}
\begin{center}
\begin{tabular}{cccccc}
\hline
\multicolumn{1}{c}{\bf Method} & \multicolumn{1}{c}{\bf BLEU (avg)}&\multicolumn{1}{c}{\bf E$\rightarrow$X}   &\multicolumn{1}{c}{\bf E$\rightarrow$X (low)}   & \multicolumn{1}{c}{\bf{X$\rightarrow$E}} & \multicolumn{1}{c}{\bf{X$\rightarrow$E (low)}} \\
\hline
 Baseline & $28.63$ &$23.01$  & $22.15$ & $34.26$ & $33.89$ \\
 \method{}& $\mathbf{29.22}$ & \bm{$23.86$}  & \bm{$22.87$} & \bm{$34.59$} & \bm{$34.34$}\\
\begin{tabular}{@{}c@{}}\extensiontwo{}\end{tabular} & $28.85$ & $23.22$ & $22.61$ & $34.49$ & $34.22$\\
\hline
\end{tabular}
\end{center}
\end{table*}

\subsection{Results on WMT-10 Dataset}
We train MoE models based on the Transformer-base architecture~\cite{vaswani2017attention} on the WMT-10 dataset for more than $100$k steps using different methods. The results are shown in Table~\ref{table:small_scale_results}.
We see that both \method{} and \extension{} have higher throughput than the baseline method with
relative improvement $10.85\%$ and $16.28\%$ respectively.
\extension{} has higher throughput than \method{} because computational cost is further reduced by skipping experts. 

As for the generalization performance, both \method{}  and \extension{} reach BLEU scores that are greater than that of the baseline when the training is converged. 
The BLEU score improvement is $0.46$ for \method{} and $0.52$ for \extension{}. 
The slightly better performance of \extension{} is likely to come from the additional regularization effect of randomly skipping expert layers.
From Table~\ref{table:small_scale_results}, we also observe that \method{} and \extension{} have faster convergence speed than the baseline. 
We choose the target BLEU score as the score achieved by the baseline method at convergence (i.e., $24.60$).
The amount of training time to reach the target BLEU score is also significantly shorter for both \method{} (by $40.89\%$) and \extension{} (by $58.48\%$) compared to the baseline.
This can also be observed in Figure~\ref{fig:small_bleu_curve}, where we show the curves of BLEU score vs. training time. 
Similarly, both \method{} and \extension{} require smaller number of training steps to reach the target BLEU score compared to the baseline.
These results confirm that our methods' fast convergence speed in terms of both the training time and the number of steps.

In addition, \citet{roller2021hash} suggest replacing the parameterized gating network with some hash function, which can alleviate the co-training issue between the gating network and experts.
As shown in Table~\ref{table:small_scale_results}, we also compare against their method Hash-Layer based on a random hash function which is simple to implement but has been shown to have very strong performance~\cite{roller2021hash}.
Hash-Layer has slightly better throughput than the baseline, which is expected because invoking a hash function is more computational efficient than passing tokens through a gating network.
However, Hash-Layer still needs the all-to-all communication to send tokens to their assigned experts (which are determined by the hash function), which explains why both \method{} and \extension{} still achieve higher throughput than Hash-Layer.
We also observe that Hash-Layer does not perform as well as our methods in terms of final BLEU at convergence.

\begin{figure}
    \centering
    \includegraphics[width=7 cm]{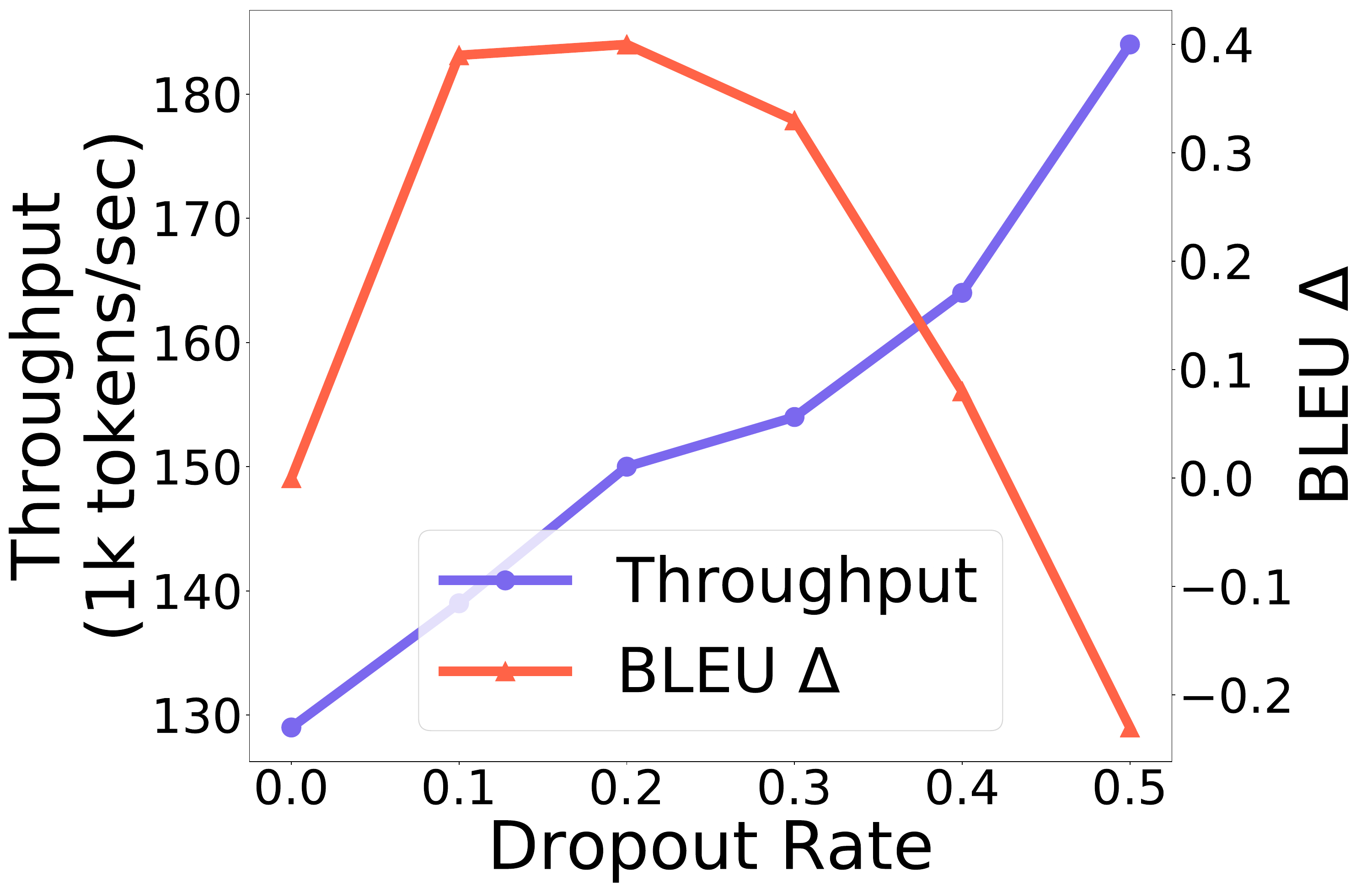}
    \caption{The effect of the dropout rate on throughput and BLEU $\Delta$ (i.e., difference w.r.t. the baseline's BLEU) for \extension{} on the WMT-10 dataset (measured after training for $35$k steps).}
    \label{fig:ablation_dropout_rate_throughput}
\end{figure}

\subsection{Results on Web-50 Dataset}
We train MoE models based on the Transformer-big architecture~\cite{vaswani2017attention} with 10 billion parameters  on the Web-50 dataset for up to $20$k steps,  it is very time consuming to train such large models for too many steps. 
In Table~\ref{table:large_scale_throughput}, we report the throughput of different methods measure in two different clusters: the cluster of V100 GPUs with a $100$Gb/s InfiniBand fabric and the cluster of A100 GPUs with a $1.6$Tb/s InfiniBand fabric. 
We observe that both \method{} and \extension{} get higher throughput than the baseline in either cluster.
It is noted that the relative throughput improvement of our methods in the V100 cluster is more significant because the hardware (GPU for computation and InfiniBand for communication) is  much slower than the A100 cluster.
Higher improvement is expected  with slower hardware where communication is the bottleneck. 

We report the best BLEU scores within $20$k steps in Table~\ref{table:large_scale_bleu}.
We can see that both \method{} and \extension{} achieve larger BLEU scores than the baseline with the improvement $0.59$ for \method{} and $0.22$ for \extension{}.
Different from the results on the WMT-10 dataset (which is a much smaller dataset), \method{} is better than \extension{} in terms of BLEU scores. 
We hypothesize that this is probably because \extension{} imposes too much regularization during training, limiting the model's fitting ability on such large scale dataset.
We further investigate the results more by looking into the BLEU score for each translation direction, i.e., from English to other languages (E$\rightarrow$X) and from other languages to English (X$\rightarrow$E). 
Especially, we consider the performance on only low-resource languages, denoted with \textbf{(low)} in Table~\ref{table:large_scale_bleu}.
Low-resource languages are those with few training data, such as Galician (Gl), Burmese (My), and Tagalog (Tl). 
It is difficult to achieve high quality translation performance on these low-resource languages.
We see that both \method{} and \extension{} improve upon the baseline on the low-resource language translation tasks by at least $0.3$ increase in BLEU score (from \extension{} on {\bf X$\rightarrow$E (low)}). 
And \method{} can yield more than $0.7$ BLEU score increase on {\bf E$\rightarrow$X (low)}).
These results suggest that our methods  improve the model's translation quality, especially on low-resource languages which would really benefit from the regularization effect of Gating Dropout.



\subsection{Effect of Dropout Rate}\label{sec:dropout_rate}
We investigate the effect of the dropout rate $p$ of \extension{} on throughput and BLEU score using the WMT-10 dataset. 
Similar observations are expected for \method{}, but we only report results for \extension{} because it outperforms \method{} on the this dataset.
We change the dropout rate from $0$ to $0.5$ for \extension{}.
When dropout rate is $0$, it is the same as the baseline method, because Gating Dropout will never be turned on in this case.
From Figure~\ref{fig:ablation_dropout_rate_throughput}, we see that, as we increase the dropout rate, the throughput also increases. 
This trend is expected because the dropout rate essentially controls how often the all-to-all communication (and the expert layer) is skipped during training.  
Larger dropout rate implies more skipping, thus resulting in higher throughput.
However, when the dropout rate is too large, the BLEU score might become worse, as shown in Figure~\ref{fig:ablation_dropout_rate_throughput}.
Specifically, when the dropout rate is $0.5$, BLEU score becomes even lower than the baseline (BLEU $\Delta$ is $-0.23$).
Actually, the BLEU score is highest when the dropout rate is $0.2$, and becomes smaller and smaller as we further increase the rate beyond $0.2$.
Similar experiments indicate that \method{} achieves highest BLEU score when dropout rate is $0.3$.
This indicates that the dropout rate needs to be properly chosen in order for our methods to work as expected.

\section{Related Work}
Several  techniques have been proposed to improve the efficacy of  MoE models. 
A new routing network which takes as input both the shared token representation and the hierarchical representation from different MoE layers is proposed in~\citet{you2021speechmoe} for speech recognition.
Expert prototyping proposed in~\citet{yang2021exploring} is shown to improve model quality by splitting experts into different prototypes, while \citet{zuo2021taming} suggest to randomly select experts to encourage learning from other experts as teachers.
Linear assignment is used to alleviate the load imbalance issue among experts~\cite{lewis2021base}.
As an effective way to regularize model training, the dropout idea has been widely studied for transformer models in general ~\cite{arora2021dropout,gao2019demystifying,kingma2015variational,boluki2020learnable}. Several dropout variants have been proposed in different settings, such as BPE-Dropout~\cite{provilkov2019bpe}, R-Drop~\cite{wu2021r}, MC-Dropout~\cite{gal2016dropout}, recurrent dropout~\cite{semeniuta2016recurrent}, curriculum dropout~\cite{morerio2017curriculum} and so on.
Unlike these dropout variants, our method  considers the communication cost which is crucial in the distributed training setting.


\section{Conclusion}
We propose Gating Dropout as a communication-efficient regularization technique to train sparsely activated transformers.
We observe that sparsely activated transformers, such as MoE models, typically have very high cross-machine communication cost, since they need to send tokens to their assigned experts via the all-to-all communication operations. 
Gating Dropout reduces the communication cost by randomly skipping the all-to-all operations.
This random skipping also has a regularization effect during training, leading to improved generalization performance. 
Experiments on multilingual translation tasks demonstrate the effectiveness of the proposed method in terms of throughput, generalization performance and convergence speed.

As for future work, we are looking at how to improve the inference speed by possibly combining Gating Dropout with expert pruning. Gating Dropout currently has no effect on the inference speed, since it is simply turned off during inference. In addition, we are also interested in the effect of varying dropout rate throughout the training process  because exploration might be much more important at the early stage of training from the exploration-exploitation perspective.   

\section*{Acknowledgements}
As part of Microsoft's effort led by Azure AI and Project Turing to build powerful large-scale language models, Gating Dropout has been utilized to train Z-Code MoE models which are adopted by Microsoft Translator\footnote{\url{https://translator.microsoft.com/}} to provide the multilingual translation service to customers.
Especially, the authors would like to thank the Z-Code team and DeepSpeed team at Microsoft for their support and feedback.

\bibliography{moe}
\bibliographystyle{icml2022}


\end{document}